\pdfoutput=1

\documentclass[11pt]{article}

\usepackage[]{emnlp2021}

\usepackage{times}
\usepackage{latexsym}
\usepackage{multirow}
\usepackage{amsmath}
\usepackage{amsfonts}
\usepackage{booktabs}
\usepackage{bm}
\usepackage{diagbox}
\usepackage{float}
\usepackage{fixltx2e}
\usepackage{todonotes}

\usepackage[T1]{fontenc}

\usepackage[utf8]{inputenc}

\usepackage{microtype}

\newcommand{\MR}[3]{\multirow{#1}{#2}{#3}}

\newcommand{\I}{\textit}
\newcommand{\T}{\texttt}

\newcommand{\true}[1]{\textbf{#1}}
\newcommand{\false}[1]{\underline{{#1}}}
\newcommand{\red}[1]{\textcolor{red}{#1}}

\title{Visual Cues and Error Correction for Translation Robustness}

\author{Zhenhao Li, Marek Rei, Lucia Specia \\
  Language and Multimodal AI (LAMA) Lab, Imperial College London \\
  \texttt{\{zhenhao.li18, marek.rei, l.specia\}@imperial.ac.uk}}

\begin{document}
\maketitle
\begin{abstract}
Neural Machine Translation models are sensitive to noise in the input texts, such as misspelled words and ungrammatical constructions. Existing robustness techniques generally fail when faced with unseen types of noise and their performance degrades on clean texts. In this paper, we focus on three types of realistic noise that are commonly generated by humans and introduce the idea of {\em visual context} to improve translation robustness for noisy texts. In addition, we describe a novel \textit{error correction training} regime that can be used as an auxiliary task to further improve translation robustness. Experiments on English-French and English-German translation show that both multimodal and error correction components improve model robustness to noisy texts, while still retaining translation quality on clean texts.
\end{abstract}

\section{Introduction}
Neural Machine Translation (NMT) has been shown to be very sensitive to noise \citep{belinkov2018synthetic,michel-neubig-2018-mtnt,ebrahimi2018adversarial}, with even small perturbations in the inputs often leading to mistranslations. To improve the robustness of NMT models, current research mostly focuses on adapting the model to noisy texts via methods such as fine-tuning \citep{michel-neubig-2018-mtnt,alam-anastasopoulos-2020-fine}, noise-injection \cite{belinkov2018synthetic,cheng-etal-2018-towards,karpukhin-etal-2019-training}, and data augmentation through back-translation \cite{berard-etal-2019-naver,vaibhav-etal-2019-improving,li-specia-2019-improving}, etc. In these approaches, the translation model is trained or fine-tuned on the noisy data so that it can learn from the noise. However, methods using extra context to help translate noisy texts have not been investigated. 

\begin{figure}[hbt]
\centering
\includegraphics[width=.75\columnwidth]{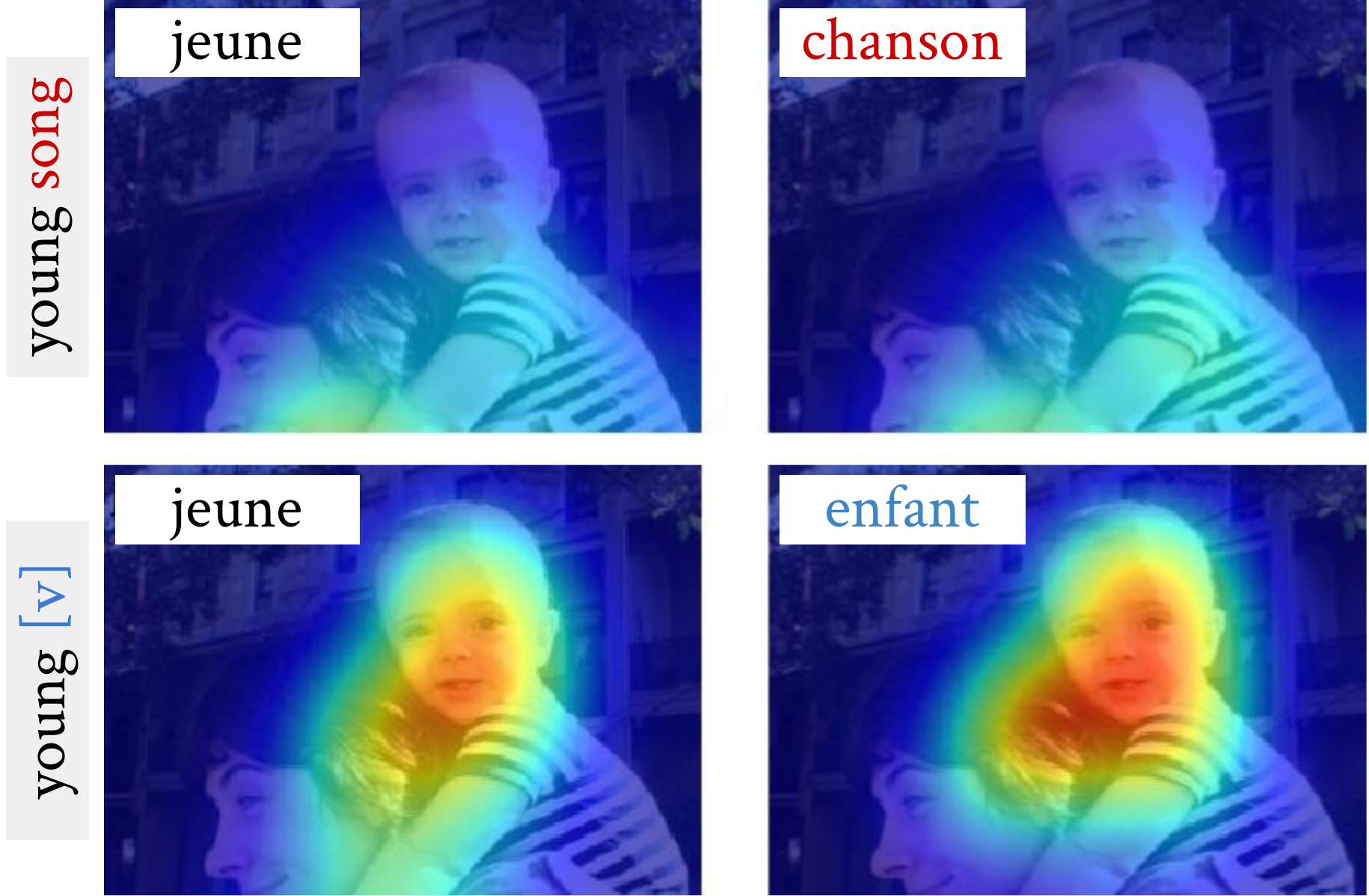}
\caption{As showed by \citet{caglayan-etal-2019-probing}, multimodality can help translate unknown words, but fail when there is noise in the input. The misspelled word ``song'' is correctly translated as ``enfant'' (child) when it is replaced with an unknown token, but translated literally as ``chanson'' (song) otherwise.} 
\label{fig:ex_att}
\end{figure}
Studies in Multimodal Machine Translation (MMT) have shown that visual information improves translation quality when the textual context is incomplete \citep{caglayan-etal-2019-probing,imankulova-EtAl:2020:WMT,caglayan-etal-2020-simultaneous}. However, as exemplified by \citet{caglayan-etal-2019-probing} (Figure \ref{fig:ex_att}), an MMT model trained on clean data was not able to handle noise. When the word ``son'' was misspelled as ``song'', the model disregarded the visual information and used the literal translation ``chanson''. The MMT model attended to the relevant region in the image and generated the intended translation ``enfant'' only when the noise was masked by a placeholder in the input, imitating an out-of-vocabulary (OOV) example. 

Given that the visual modality has been shown to help predict unknown words, 
we investigate whether adding multimodal information to adaption-based methods would further improve translation robustness. To answer this question, we build MMT models in conjunction with noise injection techniques and investigate their behaviour during training and inference on both noisy and clean data. To further improve robustness, we extend the current \textit{adversarial training} method (i.e., training NMT models on noisy texts) and propose an \textit{error correction training} method. In addition to training the model with noise-injected source sentences and their clean translation counterparts, we introduce error correction as an auxiliary task and add a separate decoder to the model, which is used to denoise the source sentence.\footnote{Codes are available at \url{https://github.com/Nickeilf/Visual-Cues-Error-Correction}} Our {\bf main contributions} can be summarized as:
\begin{itemize}
    \item To the best of our knowledge, this is the first work 
    combining adversarial training with \textit{multimodal} NMT to improve translation robustness. We evaluate robustness on three types of noise that mimic errors commonly introduced by humans. Systematic experiments reveal that multimodality can improve model performance on both known and unseen noise.
    \item We propose an \textit{error correction training} method for translation by introducing denoising as an auxiliary task. We show that the robustness of both NMT and MMT models is improved with this method.
    \item We demonstrate that the model using visual features also learns to correct grammatical errors more accurately, indicating the potential for multimodal monolingual error correction.
\end{itemize}


The paper is organised as follows: In Section \ref{sec:bg}, we present the background and related work. In Section \ref{sec:method}, we introduce the types of noise injected and the error correction training method. In Section \ref{sec:exp}, we describe our experiment settings, with experiment results in Section \ref{sec:result}, and further analysis in Section \ref{sec:analysis}. 

\section{Background and Related Work}
\label{sec:bg}
\paragraph{Robust NMT}
Although NMT models can achieve high performance on clean data, they are very brittle to non-standard inputs, such as noisy texts \citep{belinkov2018synthetic}. Different types of noisy data have been proposed to test translation robustness, e.g. synthetic word perturbations \citep{belinkov2018synthetic}, grammatical errors \citep{anastasopoulos-etal-2019-neural}, and user-generated texts from social platform \cite{michel-neubig-2018-mtnt,li-EtAl:2019:WMT1,specia-EtAl:2020:WMT1}.

The most common approach to improve translation robustness is to train the model on noisy data, which is referred to as adversarial training. Since parallel data with noisy source sentences and clean translations is difficult to obtain, the clean training data is often injected with different types of artificial noise, e.g. random word perturbations like character insertion/deletion/substitution \citep{belinkov2018synthetic,karpukhin-etal-2019-training,passban2020revisiting,xu2021addressing}, noise generated via back-translation \citep{berard-etal-2019-naver,vaibhav-etal-2019-improving,li-specia-2019-improving}, and adversarial examples generated by white-box generator model
\citep{cheng-etal-2018-towards,cheng-etal-2019-robust,cheng-etal-2020-advaug}. Even though this method has been shown to improve NMT performance on noisy data, the types of noise used thus far are not common in real data. For example, it would be highly unlikely for human authors to misspell the word ``robust'' as ``zobust'', but such random transformations are used when synthesizing noisy training data for MT. In addition, back-translation paraphrases the texts to introduce noise, however such noise is less realistic as human-generated errors, which include mispellings and grammatical errors. In adversarial approaches for other NLP tasks, \citet{ribeiro-etal-2020-beyond} and \citet{ma2019nlpaug} introduce various methods to inject both artificial and realistic noise. Inspired by these work, we focus on three types of noise that are commonly generated by humans in real texts and experiment with these for the translation task.



\paragraph{MMT} Multimodal machine translation extends the framework of NMT by incorporating extra modalities, e.g. image \citep{specia-etal-2016-shared} or audio \citep{sulubacak2020multimodal}. In our case, the extra modality is given as visual features from an image network to complement the textual context. In standard MMT, these features can be fused with the textual representation by simple operations such as concatenation \citep{caglayan2016multimodal}, hidden states initialization \citep{calixto-liu-2017-incorporating}, or via attention mechanisms \citep{libovicky-helcl-2017-attention,calixto-etal-2016-dcu,calixto-etal-2017-doubly,yao-wan-2020-multimodal} and latent variables \citep{calixto-etal-2019-latent}.

Recent research has shown that the extra modality helps translation, especially when the input is incomplete \citep{caglayan-etal-2019-probing,caglayan-etal-2020-simultaneous,imankulova-EtAl:2020:WMT} or ambiguous \citep{ive-etal-2019-distilling,DBLP:journals/corr/abs-1908-01665}. \citet{wu2019transformerbased} hinted at the possibility of multimodality helping NMT in dealing with natural noise stemming from the speech recognition system used as a first step in their pipeline approach to speech translations from videos. Their results, however, were inconclusive.

\begin{table*}[htb]
  \centering
  \begin{tabular}{ll}
    \toprule
        \textbf{clean} & a pink flower is starting to bloom .\\
        \textbf{edit-distance} & a pink flower is staring to \red{loom} .\\
        \textbf{homophone} & a pink \red{flour} is starting to bloom .\\
        \textbf{keyboard} & a pink flower is \red{starring} to bloom .\\
    \toprule
  \end{tabular}
  \caption{An example of noise injected to the clean text. The noisy substitutes are marked in red.}
  \label{tab:noise_example}
\end{table*}

\citet{salesky2021robust} investigate the robustness of open-vocabulary translation by representing texts as images followed by optical character recognition to  cover some cases of noise such as misspellings. This is an interesting but orthogonal area of research since no external visual information is used.

Therefore, it remains an open question whether MMT can perform better than NMT on noisy texts, and whether multimodality can be complementary rather than redundant to previous text-based robustness techniques. The work by \citet{caglayan-etal-2019-probing} is the closest to our approach, however they focused mainly on identifying \textit{when} the visual information is helpful. As such, they only performed experiments comparing NMT and MMT in the presence of unknown words consisting of placeholders used to mask
out words in the source sentence. In contrast, we focus on multimodal models for realistic noise that includes in- and out-of-vocabulary words, such as misspellings or correctly-spelled words used in an incorrect context. 


\section{Methods}
\label{sec:method}
In this section, we introduce our methods to improve and evaluate the robustness of NMT and MMT models. In Section \ref{sec:noise_injection}, we describe three techniques to inject realistic noise into training and test data. In Section \ref{sec:error_correction}, we introduce our error correction training method. 

\subsection{Noise Injection}
\label{sec:noise_injection}
In previous work on noise injection, the perturbations are often arbitrary, which would result in unrealistic noise. To simulate the natural noise in real situations, we add constraints to the random perturbations. We select three constrained noise injection methods that can be applied to both training and test data, with each method simulating one type of human-generated errors: 
\paragraph{Edit distance} A word is randomly replaced with another word in the vocabulary where the edit distance between the two words is less than two characters. The edit-distance noise simulates the occurrence of confusable spellings (e.g. sat vs seat) and also some grammatical errors (e.g. horse vs horses).
\paragraph{Homophones} A word is randomly replaced with another word that shares the same pronunciation. We use the CMU Pronouncing Dictionary\footnote{\url{http://www.speech.cs.cmu.edu/cgi-bin/cmudict?in=C+M+U+Dictionary}} to transform words into phonemes and find noisy substitutes with the same pronunciation. This simulates errors made by applications such as automatic speech recognition, or by non-native speakers.

\paragraph{Keyboard \citep{belinkov2018synthetic}} A character in a word is randomly replaced with an adjacent key on the standard QWERTY keyboard. The keyboard noise simulates the real-life typos when users accidentally press wrong keys while typing.
\\\\
Table \ref{tab:noise_example} shows examples of the three types of noise we experimented with. The edit distance and homophone noise types are applied on the word level, while the keyboard noise is on the character level.  
Word-level noise is more likely to break the sentence context even though the noisy substitutes are correctly spelled words. On the contrary, character-level noise is likely to introduce misspelled words and increase the out-of-vocabulary (OOV) rate.

When constructing the noisy training or test sets, we sample from the three types of noise following a uniform distribution, where to each sentence we apply only one type of noise. To avoid substituting words not carrying much contextual information (e.g. articles and punctuations)
, we only perturb words with more than two characters. The noise level is controlled by the hyperparameter $\mathit{n}$, which defines the maximum number of words replaced with noisy counterparts per sentence. The noise injection procedure can be characterized as: given a source sentence $\mathbf{x} = [x_1, x_2, ..., x_M]$ and a target translation $\mathbf{y} = [y_1, y_2, ..., y_N]$, noise will be injected to the clean source sentence $\mathbf{x}$ to obtain its noisy variant $\mathbf{x'} = [x_1, ..., x'_{a_i} ,..., x_M]$, where $a_i$ is the position of the noisy substitutes ($i=\{1,2,...,n\}$).

\begin{figure*}
\centering
  \includegraphics[width=0.35\textwidth]{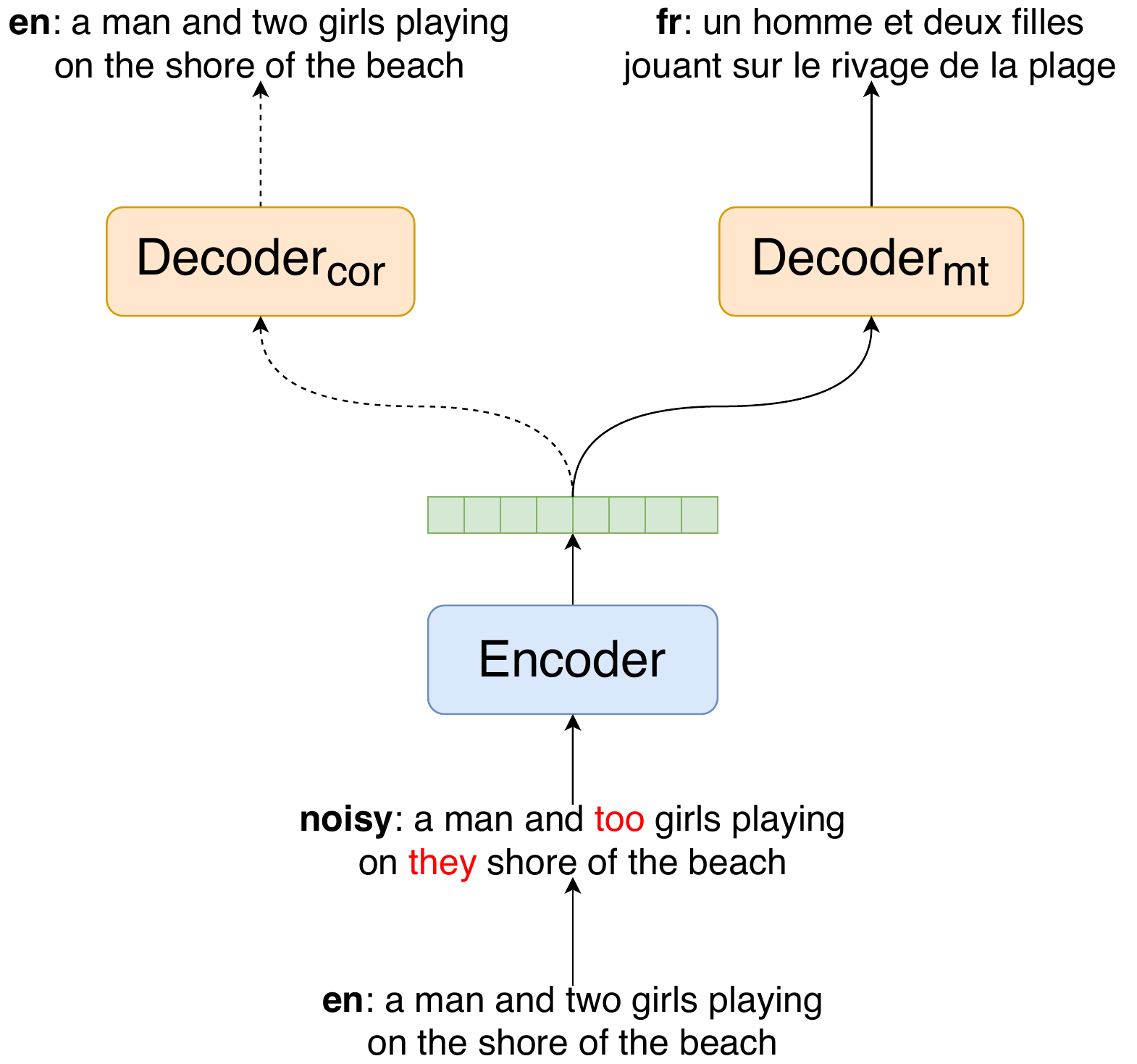}
  \includegraphics[width=0.334\textwidth]{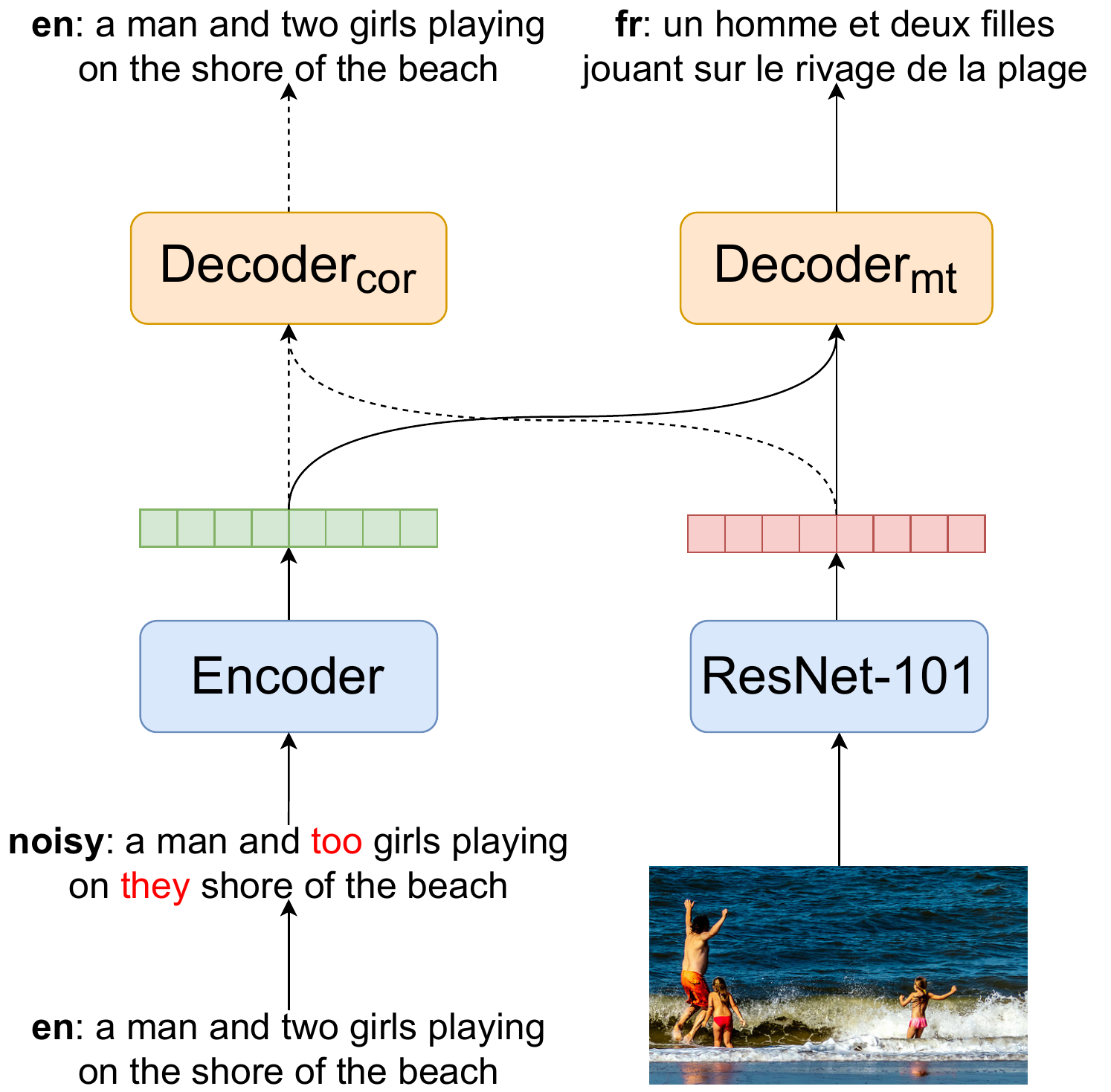}
  \caption{Illustration of the joint training of machine translation and error correction for NMT and MMT models. Solid lines: translation flow. Dotted lines: error correction flow. Left: NMT with error correction training. Right: MMT with error correction training.}
  \label{fig:model}
\end{figure*}

\subsection{Error Correction Training}
\label{sec:error_correction}
We introduce error correction \citep{ng-etal-2014-conll,yuan-briscoe-2016-grammatical} as an auxiliary task to help improve the robustness against noisy inputs. For that, we add a second decoder to the MT architecture, which is only used for the error correction task.
During training, the noisy sentence $\mathbf{x'}$ is encoded by the encoder, which is shared between the translation and correction tasks, into hidden states $\mathbf{h'}$. The hidden state representation is then fed to both decoders. The translation decoder aims to generate a correct translation $\mathbf{y}$ while the correction decoder aims to recover the original source sentence $\mathbf{x}$. This method is also compatible with the MMT model, where the error correction decoder will use both visual and textual hidden states to recover the clean source sentences. Figure \ref{fig:model} gives an illustration of the model architecture. 

Compared to the standard MT model, the version with error correction training (which we refer to as \textit{NMT-cor} and \textit{MMT-cor} hereinafter) maximizes both the probability of generating correct translations $P(\mathbf{y}|\mathbf{x'};\bm{\theta_{mt}})$ and the probability of recovering the clean source sentences $P(\mathbf{x}|\mathbf{x'};\bm{\theta_{cor}})$. 
\begin{equation}
\begin{aligned}
    P(\mathbf{y}|\mathbf{x'};\bm{\theta_{mt}}) &= \prod_{t=1}^{N}P(y_t|\mathbf{y}_{1:t-1},\mathbf{x'};\bm{\theta_{mt}}) \\
    P(\mathbf{x}|\mathbf{x'};\bm{\theta_{cor}}) &= \prod_{t=1}^{M}P(x_t|\mathbf{x}_{1:t-1},\mathbf{x'};\bm{\theta_{cor}}) \\
\end{aligned}
\end{equation}
The $\bm{\theta_{mt}}$ represents parameters for the translation component and the $\bm{\theta_{cor}}$ represents parameters for the error correction component, with $\bm{\theta_{mt}} = \{\bm{\theta_{enc}}, \bm{\theta_{mt\_dec}}\}, \bm{\theta_{cor}} = \{\bm{\theta_{enc}}, \bm{\theta_{cor\_dec}}\}$. Our hypothesis is that the auxiliary task of error correction may help the encoder with a noise-invariant representation, which would indirectly improve the translation of noisy sentences. During training, we jointly optimize the sum of the translation loss and the error correction loss, as is shown in Equation \ref{eq:loss}:
\begin{equation}
\label{eq:loss}
\begin{aligned}
    \mathcal{L}_{mt}(\boldsymbol{\theta}_{mt}) &= \frac{1}{|\mathbf{D}|}\sum_{(\mathbf{x'},\mathbf{y})\in\mathbf{D}}-\log P(\mathbf{y}|\mathbf{x'};\boldsymbol{\theta}_{mt}) \\
    \mathcal{L}_{cor}(\boldsymbol{\theta}_{cor}) &= \frac{1}{|\mathbf{D}|}\sum_{(\mathbf{x'},\mathbf{x})\in\mathbf{D}}-\log P(\mathbf{x}|\mathbf{x'};\boldsymbol{\theta}_{cor}) \\
    \mathcal{L}(\boldsymbol{\theta}) &= \mathcal{L}_{mt}(\boldsymbol{\theta}_{mt}) + \boldsymbol{\lambda}\mathcal{L}_{cor}(\boldsymbol{\theta}_{cor})
\end{aligned}
\end{equation}
where $\boldsymbol{\lambda} \ge 0$ is the factor that controls the weight of the error correction loss, and $\mathbf{D}$ represents the noise-injected data consisting of triples in the form of $(\mathbf{x}, \mathbf{x'}, \mathbf{y})$.

\section{Experiments}
\label{sec:exp}
\subsection{Datasets}
We experiment with the Multi30K dataset \citep{elliott-etal-2016-multi30k}, using both the En-Fr and En-De language pairs. This is the standard dataset for MMT and has been used in all open challenges on the topic \citep{specia-etal_WMT:2016,elliott-EtAl:2017:WMT,barrault-etal_WMT:2018}. Following \citet{caglayan-etal-2019-probing}, we use both the \textit{train} and \textit{valid} splits as our training set. The \textit{test2016-flickr} set is used as our development set for checkpoint selection. For evaluation, we test the models on both \textit{test2017-flickr} and \textit{test2017-mscoco} sets \citep{elliott-etal-2017-findings}. We use a word-level vocabulary and build vocabularies for the original source and target languages, as well as the vocabulary on noisy source texts.\footnote{Therefore there is no OOV word in the noisy training data, but the test data might still contain OOV words -- noisy or not -- with respect to the training.} We use the pre-processed data in Multi30K, which is lowercased, normalized, and tokenized with Moses \cite{koehn-etal-2007-moses}. We also performed experiments using a subword-level vocabulary (BPE), which led to further improvements, but the trend in the results is the same (see Appendix \ref{sec:subword}). 

Following \citet{caglayan-etal-2020-simultaneous}, we use the ``bottom-up-top-down'' (BUTD) features \citep{butd} extracted from a pre-trained Faster R-CNN ResNet-101 object detector. Each image is represented as 36 pooled feature vectors $V \in \mathbb{R}^{36\times2048}$, with each vector representing a local object region.

\subsection{Models}
\paragraph{NMT and MMT Models}
Our baseline NMT model is the standard Transformer model \citep{attention-is-all-you-need},
with 6 layers for both the encoder and the decoder. The hidden state size is 512 while the feed-forward dimension is 1024. The number of attention heads is set to 4. Dropout (0.3) is applied to both self/cross-attention and the position-wise feed-forward layer, and Pre-norm \cite{nguyen2019transformers} is applied to boost convergence. Our baseline MMT model follows the same architecture and hyperparameters as the baseline NMT model, except for the multimodal components. We use the serial multimodal cross-attention \citep{libovicky-etal-2018-input}, where an extra cross-attention sublayer is appended in the decoder layer to perform attention over the visual features. We also experiment with GRU models \citep{cho-etal-2014-learning}, following the hyperparameter settings of \citet{caglayan-etal-2019-probing}. Due to space restrictions, we include the detailed results with GRU models in Appendix \ref{sec:gru_results}. The GRU results display the same trend as the experimental results using Transformer models.

\paragraph{Error Correction Models}
The error correction NMT/MMT models adopt the same encoder and decoder as the baseline NMT/MMT models, except for a second decoder added for error correction training. During training, we compute the cross-entropy loss for  translation, as well as for error correction in the correction-based models. In these models, the two losses are summed and optimized jointly on the same batch. 
We found the best $\lambda$ value ($\lambda \in \{0.2, 0.2, 0.4, 0.4, 0.8\}$) for different levels of noise (number of noisy words $n \in \{1, 2, 4, 6, 10 \}$) during hyperparameter tuning. See Appendix \ref{sec:lambda} for more details.

\paragraph{Training and Evaluation}
We use ADAM \cite{kingma2017adam} as the optimizer and adopt the \textit{noam} learning rate scheduler \cite{attention-is-all-you-need} with a warm-up of 8000 steps. The training batch size is 64. Models are evaluated using the METEOR score \citep{denkowski-lavie-2014-meteor}, which is the main metric for multimodal machine translation \cite{barrault-etal_WMT:2018}.
For the evaluation of error correction, we use ERRANT \citep{bryant-etal-2017-automatic} to compute the $F_{0.5}$ score. During evaluation, we select the checkpoint with the best performance on the development set and generate the translation and correction using beam search of size 12. All models are implemented using \textit{nmtpytorch}\footnote{\url{https://github.com/lium-lst/nmtpytorch}} and \textit{pysimt}\footnote{\url{https://github.com/ImperialNLP/pysimt}}. Each model is run with three random seeds and the average results are reported. Each run takes approximately 2 hours to train on an RTX 2080 Ti GPU.

\section{Results}
\label{sec:result}
\subsection{Testing for Robustness to Noise}
\begin{figure}[htb]
\centering
  \includegraphics[width=.95\linewidth]{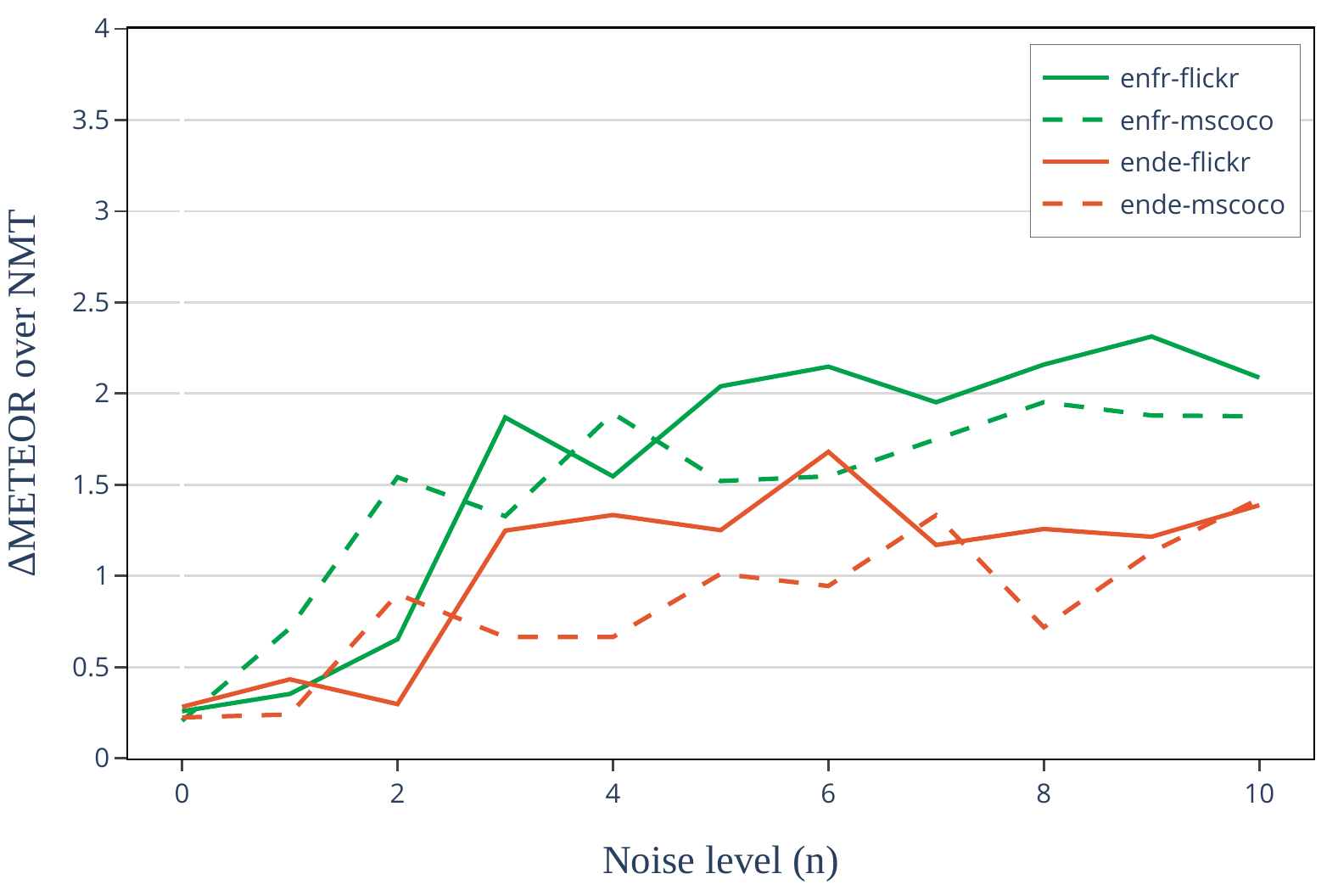}
  \caption{Performance gain from multimodality on different test sets when models are \textit{trained on clean data} but \textit{tested on noisy data} ($\Delta$METEOR = MMT - NMT).}
  \label{fig:clean_train}
\end{figure}

We first evaluate the robustness of standard NMT and MMT models \textbf{trained on clean data} by \textbf{testing on the noise-injected data}. This setting represents regular models that are not specifically adapted to noise. Figure \ref{fig:clean_train} presents the change in METEOR ($\Delta$METEOR) between standard MMT and NMT models tested on data with different noise levels. The $\Delta$METEOR is consistently above 0 for both test sets in the two language pairs. As the noise level increases, the difference between NMT and MMT models is larger, showing that the visual information in the MMT model leads to predictions that are more robust to noise.

\begin{table*}[htb]
    \centering
    \begin{tabular}{l|c c c c c c | c c c c c c}
    \toprule
     & \multicolumn{6}{c}{\textbf{flickr2017}} & \multicolumn{6}{|c}{\textbf{mscoco2017}}  \\
     & clean & $n$=1 & $n$=2 & $n$=4 & $n$=6 & $n$=10 & clean & $n$=1 & $n$=2 & $n$=4 & $n$=6 & $n$=10 \\
    \midrule
    \multicolumn{11}{@{}l}{\textit{en-fr}} \\
    
    NMT & 70.6 & 64.2 & 60.2 & 55.2 & 51.8 & 49.4 & 64.2 & 58.3 & 54.3 & 48.8 & 45.7 & 43.2 \\
    MMT & 70.9 & 64.7 & 61.0 & 56.8 & 53.7 & 51.1 & 64.4 & 59.3 & 55.4 & 50.1 & 47.8 & 45.2 \\
    NMT-cor & --- & 64.9 & 61.6 & 57.4 & 54.7 & 55.0 & --- & 59.2 & 55.2 & 51.4 & 48.0 & 47.2 \\
    MMT-cor & --- & \bf 65.2 & \bf 62.2 & \bf 59.0 & \bf 56.7 & \bf 55.5 & --- & \bf 59.6 & \bf 56.4 & \bf 52.4 & \bf 50.0 & \bf 48.9 \\
    \midrule
    \multicolumn{11}{@{}l}{\textit{en-de}} \\
    
    NMT & 52.3 & 47.2 & 44.3 & 40.2 & 38.4 & 36.7 & 47.5 & 43.5 & 40.2 & 36.8 & 34.0 & 32.5 \\
    MMT & 52.6 & 47.7 & 45.2 & 41.3 & 39.3 & 37.6 & 47.7 & 43.9 & 41.0 & 37.9 & 35.1 & 33.9 \\
    NMT-cor & --- & 47.9 & 45.6 & 42.9 & 41.4 & 41.1 & --- & \bf 44.2 & 41.9 & 38.4 & 36.8 & 36.2 \\
    MMT-cor & --- & \bf 48.0 & \bf 46.1 & \bf 43.5 & \bf 42.5 & \bf 41.8 & --- & 43.9 & \bf 42.3 & \bf 39.7 & \bf 38.2 & \bf 37.4 \\
    \bottomrule
    \end{tabular}
    \caption{Results in METEOR scores of models trained and tested on different levels of noisy data. The train and test data are injected with the same proportion of noise. $n$ indicates the max number of noisy words in the train/test set. *-cor indicates the models with error correction training.}
    \label{tab:mix_noise}
\end{table*}

\subsection{Training for Robustness to Noise}
To test models for their ability to adapt to noisy data, we \textbf{train models on data with added noise}, sampling from the three types of noise in Section~\ref{sec:noise_injection} and \textbf{test them on noisy test data}, with noise added in the same fashion. METEOR score results are shown in Table \ref{tab:mix_noise}.

The training on noisy data is equivalent to the ``adversarial training'' experiments in previous studies \citep{belinkov2018synthetic,karpukhin-etal-2019-training}. In this setting, a text-only NMT model still suffers from significant performance degradation as the number of noisy words grows, for example dropping from 70.6 METEOR on clean test data to 49.4 under the noisiest setting for en-fr on flickr2017. A drop is also observed for the MMT model, however it is smaller for both language pairs and test sets. As $n$ becomes larger, the gain from the visual context is more obvious, showing that additional context in the form of image features is increasingly important for translation when the quality of the textual input is degraded.

With the addition of the error correction training, both NMT and MMT models further improve their performance, with NMT-cor even outperforming the base MMT model. The MMT-cor model performs better than both NMT-cor and base MMT models, demonstrating that the improvements from error correction and visual cues are complementary. Similar to the benefit from visual features, the difference between models with and without error correction training becomes larger when the noise level increases.

In addition to the performance on noisy texts, another important aspect when measuring robustness is to evaluate whether the performance of the models on clean data is harmed when the model is adapted to the noisy data. Following \citet{karpukhin-etal-2019-training}, we {\bf train models on a mixture of noisy and clean data (0.5/0.5)} and {\bf test them on clean (original) data}. 
Table \ref{tab:clean_drop} shows the performance drop on the clean Flickr2017 En-Fr test set, compared to the baseline NMT model trained with clean data only. 
\begin{table}[htb]
    \centering
    \begin{tabular}{l | c c c c c}
    \toprule
         $n$ = & 1 & 2 & 4 & 6 & 10 \\
    \midrule
        
        
        NMT & $\downarrow$0.2 & $\downarrow$1.0 & $\downarrow$1.4 &	$\downarrow$2.0 & $\downarrow$2.3 \\
        MMT & $\downarrow$0.2 & $\downarrow$0.7 & $\downarrow$1.7 &	$\downarrow$2.1 & $\downarrow$2.4 \\
        MMT-cor & $\downarrow$0.0 & $\downarrow$0.4 & $\downarrow$0.9 &	$\downarrow$1.7 & $\downarrow$2.1 \\
        
    \bottomrule
    \end{tabular}
    \caption{Performance drop (the lower the better) on clean Flickr2017 En-Fr test set when models are \textit{trained on mixed data}, compared to baseline NMT model (70.6 METEOR) trained on clean data.}
    \label{tab:clean_drop}
\end{table}

The trend is same for models on the other datasets/language pairs: the larger the proportion of noise in the training data, the higher the performance drop on the clean test set. 
However, the largest drop in METEOR is only 2.4, showing that mixing clean and noisy training data is a good strategy.\footnote{In additional experiments, we found that models trained on entirely noisy data show much more severe performance drops as \textit{$n$} becomes larger -- see Appendix \ref{sec:train_noise_test_clean}.} 
Both MMT and MMT-cor show a similar performance drop to the base NMT model, which indicates that the use of visual context and error correction training does not harm performance on clean texts.

The corresponding  results for Table \ref{tab:mix_noise} and \ref{tab:clean_drop} with GRU models can be found in Appendix \ref{sec:gru_results}, showing a similar benefit when using multimodal information and error correction training.

\section{Analysis}
\label{sec:analysis}
\paragraph{Robustness on Unseen Noise}
Since in realistic applications the noise distribution at test time is unknown, we evaluate models using different noise proportions and  types at training and test time. For the former, we test the {\em same model} ($n$=4) on various test sets created with different values of $\textit{n}$. For the latter, we test the same model ($n$=4) on the test set where words are randomly replaced with unknown tokens (i.e. ``[UNK]'') to simulate unseen noise (noisy words from different corpora or domains, e.g. new emojis). 
Table~\ref{tab:unk} shows results for both cases. 

\begin{table}[htb]
    \centering
    \begin{tabular}{l | c c c c}
    \toprule
        \textit{$n$} = & 1 & 2 & 6 & 10 \\
    \midrule
        NMT & 62.5 & 59.3 & 51.6 & 49.2 \\
        MMT & 62.9 & 60.1 & 52.8 & 51.0 \\
        MMT-cor & {\bf 64.1} & {\bf 62.0}  & {\bf 55.5} & {\bf 53.8} \\
    \midrule
        UNK= & 1 & 2 & 3 & 4 \\
    \midrule
        NMT & 55.5 & 46.7 & 38.7 & 31.6 \\
        MMT & 57.0 & 48.8 & 41.2 & 34.8 \\
        MMT-cor & {\bf 57.9} & {\bf 49.9} &	{\bf 42.6} &	{\bf 36.1} \\
    \bottomrule
    \end{tabular}
    \caption{Performance of NMT and MMT models trained noisy data with $n$=4 but tested on data with different noise proportion and noise types. All models are tested on Flickr2017 En-Fr.}
    \label{tab:unk}
\end{table}

The overall trend is similar to the case when the train/test noise are the same: models with visual information and error correction training achieve better performance. The METEOR score of train/test noise proportion mismatch is close to the score in Table~\ref{tab:mix_noise} under the same noise proportion
, showing that the models are robust to unknown noise distributions. As for the evaluation on unknown noise types, the MMT model outperforms the NMT model, which indicates the better ability of the MMT model to handle unseen noise. 

\begin{table*}[htb]
\small
\renewcommand{\arraystretch}{0.95}
\centering
\begin{tabular}{cl@{}}
\toprule
\MR{10}{*}{\includegraphics[height=2.5cm]{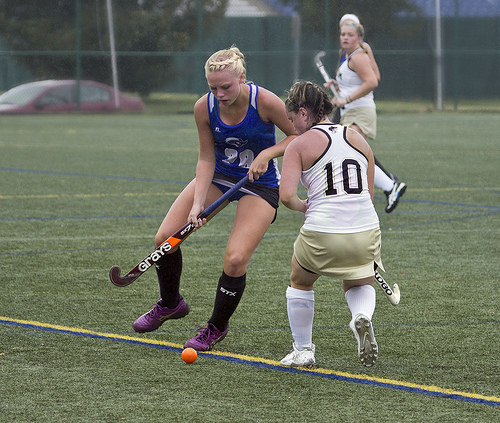}} &
\T{SRC}: women are playing lacrosse with an orange ball . \\
& \T{NSY}: women [art] playing lacrosse with an [strange] ball . \\
&\T{NMT}: des femmes jouent au lacrosse avec une balle \false{étrange} . \\
& \T{NMT}\textsubscript{cor}: des femmes jouent au lacrosse avec une boule \false{étrange} . \\
& \phantom{\T{NMT}\textsubscript{cor}:} \I{(women are playing lacrosse with a strange ball.)} \\
& \T{MMT}\textsubscript{cor}: des femmes jouent \true{à la} lacrosse avec une balle \true{orange} . \\
& \T{REF}: des femmes jouent \true{à la} crosse avec une balle \true{orange} . \\
& \phantom{\T{NMT}\textsubscript{cor}:} \I{(women are playing lacrosse with an orange ball .)} \\
& \T{COR-NMT}: women \true{are} playing lacrosse with an \false{old} ball . \\
& \T{COR-MMT}: women \true{are} playing lacrosse with an \true{orange} ball . \\
\midrule
\MR{10}{*}{\includegraphics[height=2.5cm]{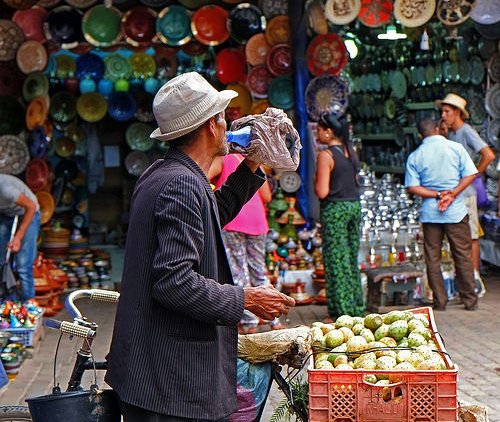}} &
\T{SRC}: a man with his bicycle selling his products on a street \\
& \T{NSY}: a [kan] with his [bicycld] selling his products on a street \\
& \T{NMT}: un homme avec son \false{casque} vendant ses produits dans une rue \\
& \phantom{\T{NMT}:} \I{(a man with his helmet selling his products on a street)} \\
& \T{NMT}\textsubscript{cor}: un homme avec son \true{vélo} vendant ses produits dans une rue \\
& \T{MMT}\textsubscript{cor}: un homme avec son \true{vélo} vendant ses produits dans une rue \\
& \T{REF:} un homme avec son \true{vélo} vendant ses produits dans une rue \\
& \phantom{\T{NMT}\textsubscript{cor}:} \I{(a man with his bicycle selling his products on a street)} \\
& \T{COR-NMT}: a man with his \true{bicycle} selling his products on a street \\
& \T{COR-MMT}: a man with his \true{bicycle} selling his products on a street \\
\bottomrule
\end{tabular}
\caption{Qualitative examples for both translation and error correction, where noise is indicated by the words in square brackets. Underlined and bold words highlight the bad and good lexical choices, respectively. \texttt{NSY}: noisy sentence. \texttt{COR-*}: corrected sentence (output from the error correction decoder).}
\label{tbl:imgcomp}
\end{table*}

\begin{figure}
    \centering
    \includegraphics[width=.95\linewidth]{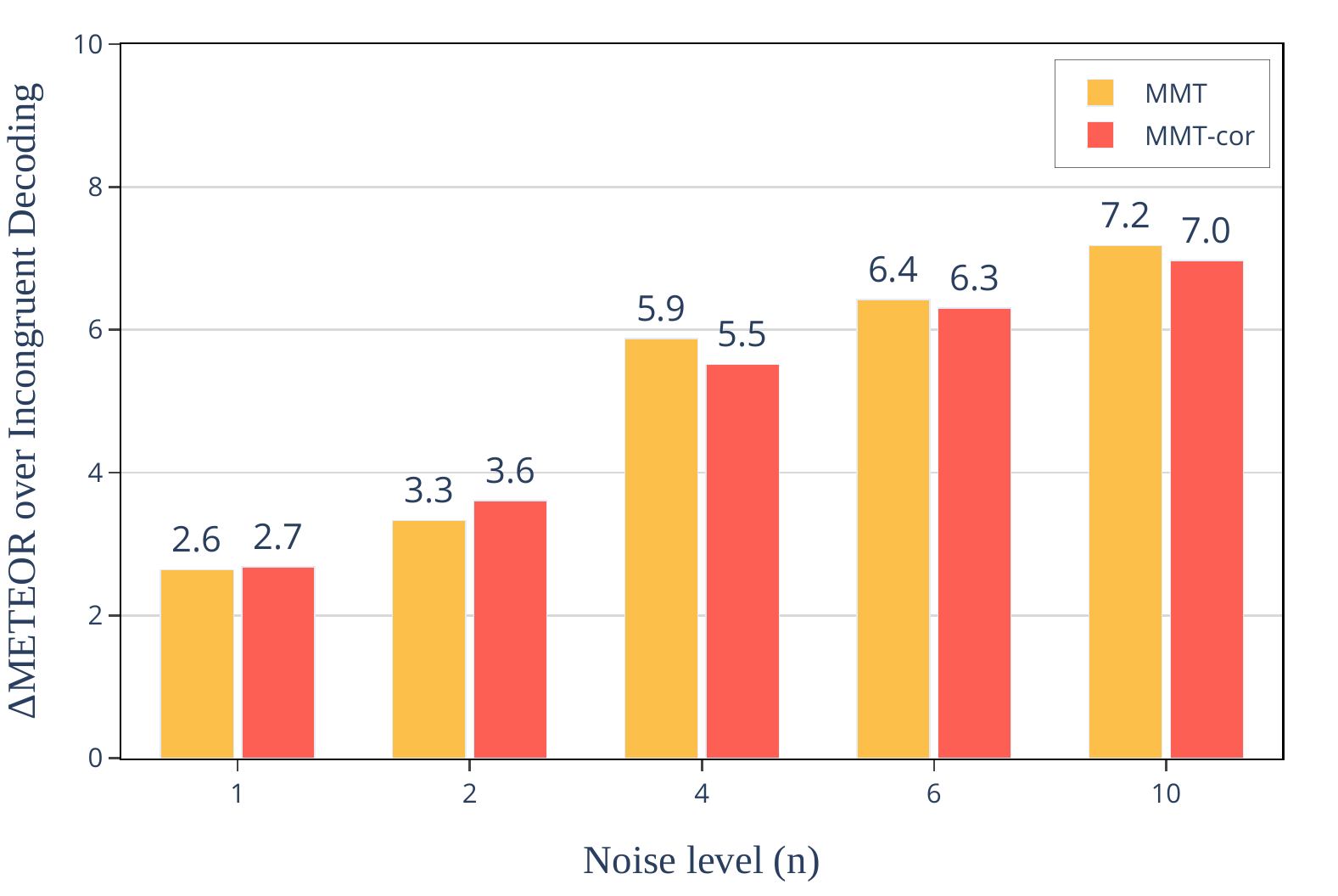}
    \caption{Performance gap in METEOR score between congruent decoding and incongruent decoding ($\Delta$METEOR = congruent - incongruent).}
    \label{fig:incongruent}
\end{figure}

\paragraph{Visual Sensitivity}
To further probe the effect of the visual information on MMT and MMT-cor models, we apply the \textit{incongruent decoding} evaluation approach \citep{elliott-2018-adversarial,caglayan-etal-2019-probing} by feeding the multimodal models with incorrect visual features at test time, i.e. features taken from a different test sample. The expectation is that the multimodal model will suffer due to the incorrect visual context, performing worse compared to using the correct visual features. Figure \ref{fig:incongruent} shows the performance gap between congruent decoding and incongruent decoding.

The $\Delta$METEOR is always positive for both MMT and MMT-cor models, and this difference is amplified with a larger noise ratio in the test data, reaching up to 7.2 METEOR scores when $n$=10. We note that the $\Delta$METEOR for the MMT-cor model is similar to the MMT model, but slightly lower, indicating that the error correction training helps the model recover from incorrect image features to a small extent on noisier data.


\paragraph{Error Correction Quality}
To understand whether visual information can also benefit error correction, we compute the span-based correction $F_{0.5}$ score as commonly used in the Grammatical Error Correction task \citep{dahlmeier-ng-2012-better}. The $<$noisy, corrected$>$ and $<$noisy, clean$>$ pairs are first transformed into two lists of edits, where adding/replacing/deleting a word at any position counts as one edit. The evaluation is then performed by calculating the precision/recall/F0.5 between these edit sets.

We report the results in Table \ref{tab:error_correction} for both NMT-cor and MMT-cor models trained on different values of \I{$n$}. The MMT-cor model outperforms the NMT-cor model, with an improvement of up to +1.7 and +2.6 $F_{0.5}$ on the two test sets. This improvement indicates that visual features can also be beneficial for error correction performance, showing a potential for the task of multimodal error correction, which has yet to be explored.


        

\begin{table}[htb]
    \centering
    \resizebox{\linewidth}{!}{
    \begin{tabular}{l | c c c | c c c}
    \toprule
        & \multicolumn{3}{c|}{flickr2017} & \multicolumn{3}{c}{mscoco2017} \\
        & Prec & Rec & $F_{0.5}$ & Prec & Rec & $F_{0.5}$ \\
    \midrule
    \multicolumn{7}{@{}l}{\textit{$n$}=1} \\
        NMT-cor & 41.9 & 52.5 & 43.7 & 45.1 &  51.4  & 46.2 \\
        MMT-cor & \bf43.3 & \bf54.0 & \bf45.1 & \bf46.5 &  \bf53.8 &  \bf47.8 \\
    \midrule
    \multicolumn{7}{@{}l}{\textit{$n$}=2} \\
        NMT-cor & 56.7 & 62.2 & 57.7 & \bf53.2 &  56.2  & \bf53.8 \\
        MMT-cor & \bf57.0 & \bf63.3 & \bf58.1 & 52.9 & \bf56.4 &  53.6 \\
    \midrule
    \multicolumn{7}{@{}l}{\textit{n=4}} \\
        NMT-cor & 66.6 & 69.1 & 67.1 & 65.7 & 66.7 & 65.9 \\
        MMT-cor & \bf67.7 & \bf71.5 & \bf68.5 & \bf66.1 & \bf67.6 & \bf66.4 \\
    \midrule
    \multicolumn{7}{@{}l}{\textit{$n$}=6} \\
        NMT-cor &  68.7 & 70.0 & 69.0 & 67.0 & 66.1 & 66.8 \\
        MMT-cor &  \bf70.4 &  \bf71.8 &  \bf70.7 & \bf68.3 &  \bf67.6 & \bf68.2\\
    \midrule
    \multicolumn{7}{@{}l}{\textit{$n$}=10} \\
        NMT-cor &  72.5 & 73.2 & 72.6 & 67.1 & 66.4 & 67.0 \\
        MMT-cor &  \bf73.9 &  \bf74.5 & \bf74.1 & \bf69.8 & \bf68.6 & \bf69.6\\
    \bottomrule
    \end{tabular}
    }
    \caption{Error Correction score in $F_{0.5}$ for both NMT-cor and MMT-cor models.}
    \label{tab:error_correction}
\end{table}

\begin{figure*}[htb]
\centering
  \includegraphics[width=\textwidth]{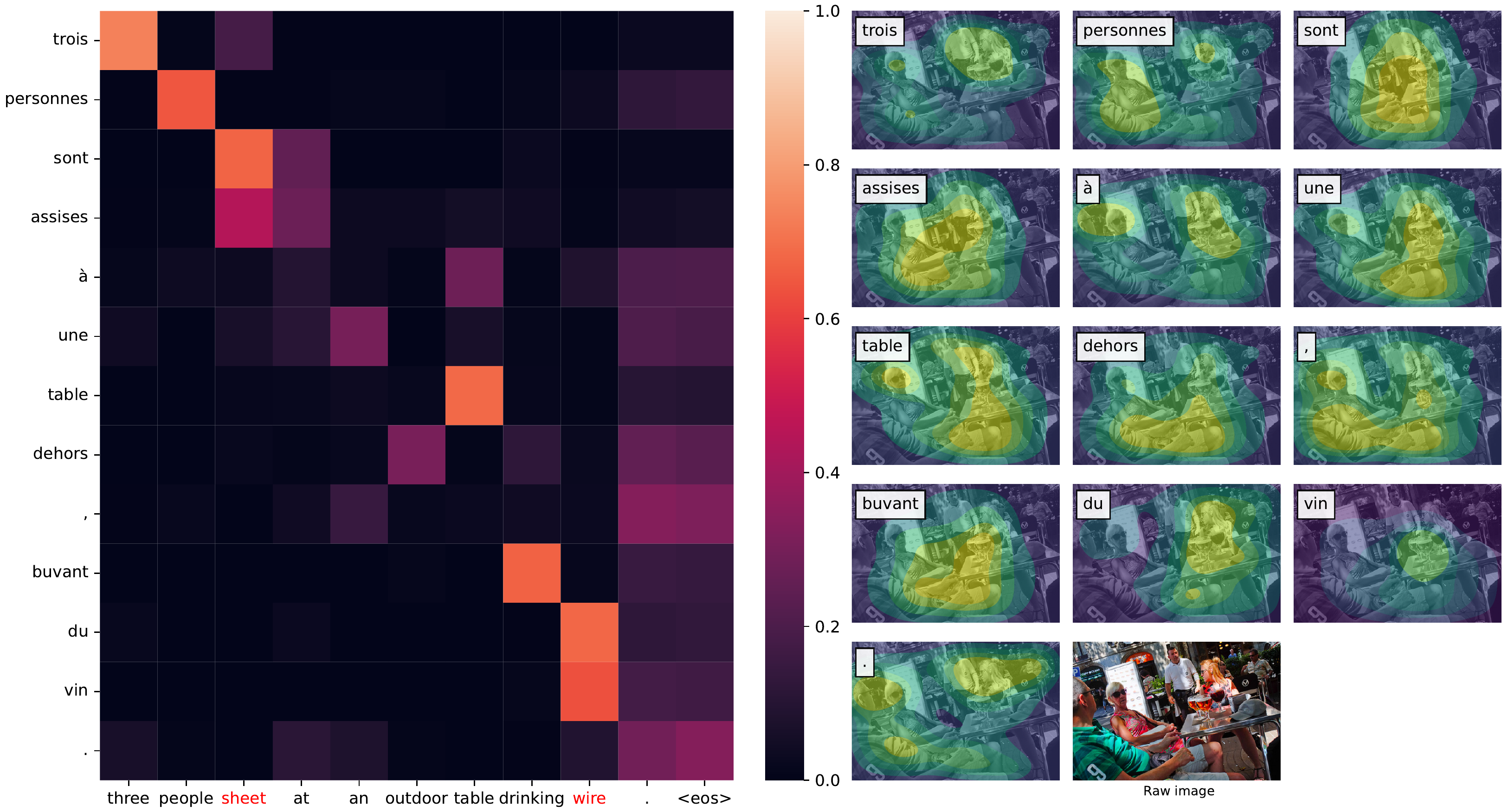}
  \caption{Attention map of the MMT-cor system on input texts and visual features when generating the translation from noisy input with the target decoder.}
  \label{fig:trans_attention}
\end{figure*}


\paragraph{Qualitative Examples}
We provide two qualitative examples of the visual features and error correction training helping the model handle input noise in Table \ref{tbl:imgcomp} 
(see Appendix \ref{sec:qualitative_examples} for more examples). In the first example, the source sentence is injected with  the ``edit-distance'' noise, with ``are'' and ``orange'' replaced with ``art'' and ``strange'' respectively. Both NMT and NMT-cor models fail to include ``orange'' in the translation, as it is difficult to recover from this error without visual information, while the MMT-cor model is able to generate the correct output. The source sentence in the second example is injected the ``keyboard'' noise, with ``man'' replaced with ``kan'' and ``bicycle'' replaced with ``bicycld''. Although the training data is injected with the same types of noise, the NMT model fails to translate correctly. The reason might be that ``bicycle'' has multiple noisy variants, such as ``bicycld'', ``bocycle'', etc., so the NMT model can hardly learn a strong relationship between ``bicycld'' and ``vélo'' (translation of ``bicycle''). However, the NMT-cor model could relate ``bicycld'' with ``bicycle'', which helps to predict the correct translation ``vélo''.

In Figure \ref{fig:trans_attention}, we also present the attention map of the MMT-cor system when generating the translation. The input is injected with noise by substituting ``sit'' with ``sheet'', and ``wine'' with ``wire''. When generating ``sont assises'' (are sitting), although the attention on the input text still mainly focuses on the noisy word ``sheet'' (with a small proportion focusing on the preposition ``at''), the visual attention is able to focus on the people in the image; therefore, the model obtains the correct information from the visual input and is able to generate the correct translation. Similarly, the model generates ``vin'' (wine) by attending to the glasses in the images and is not distracted by the noisy input word ``wire''.
The attention map for the example when generating the error correction output can be found in Figure \ref{fig:cor_attention} in the Appendix.


\section{Conclusions}
\label{sec:conclusion}
In this paper we propose to explore visual cues in order to improve model robustness to noise in machine translation. We combine adversarial training on artificially generated noisy examples with visually-informed multimodal machine translation. By training multimodal models on noisy data, we show that the extra visual context can improve translation robustness on both known and unseen noise. We also propose a novel error correction training method, jointly optimizing the translation model with an auxiliary objective for correcting input errors, which we show can further improve the robustness of both text-only and multimodal translation models. Future work in this area could investigate the integration of further modalities, such as audio in the speech translation setting. In addition to translation, we found that the model using visual features can also help correct errors in the source language. This opens up a promising direction for multimodal monolingual error correction, a task not yet explored.

\section*{Acknowledgements}
Lucia Specia and Zhenhao Li received support from MultiMT project (H2020 ERC Starting Grant No. 678017) and the Air Force Office of Scientific Research (under award number FA8655-20-1-7006).

\bibliography{anthology,custom}
\bibliographystyle{acl_natbib}

\clearpage
\appendix
\section{Word-level vs. Subword-level}
\label{sec:subword}
In Table \ref{tab:subword} we present the results for NMT and MMT models using word-level and subword-level vocabulary. Models using byte-pair-encoding (BPE) perform better than models with word-level vocabulary. Nevertheless, MMT models ourperform NMT models when using BPE. Likewise, the MMT-cor models are consistently better than the MMT model when subword-level vocabulary is applied. The results show that the benefit from both multimodality and error correction training still holds on models with subword-level vocabulary.
\begin{table}[htb]
    \centering
    \resizebox{\linewidth}{!}{
    \begin{tabular}{l c c c c c c }
    \toprule
     & \multicolumn{6}{c}{\textbf{flickr2017 En-Fr}} \\
     & clean & $n$=1 & $n$=2 & $n$=4 & $n$=6 & $n$=10  \\
    \midrule
    \multicolumn{6}{@{}l}{\textit{w2w}} \\
    NMT & 70.6 & 64.2 & 60.2 & 55.2 & 51.8 & 49.4 \\
    MMT & 70.9 & 64.7 & 61.0 & 56.8 & 53.7 & 51.1 \\
    MMT-cor & --- & 65.2 & 62.2 & 59.0 & 56.7 & 55.5 \\
    \midrule
    \multicolumn{6}{@{}l}{\textit{bpe2w}} \\
    NMT & 70.5 & 65.2 & 61.4 & 56.4 & 53.6 & 51.5 \\
    MMT & 70.8 & 65.6 & 62.1 & 58.0 & 54.9 & 53.1 \\
    MMT-cor & --- & 65.9 & 63.6 & 60.8 & 58.3 & 57.2 \\
    \midrule
    \multicolumn{6}{@{}l}{\textit{bpe2bpe}} \\
    NMT & 70.8 & 65.5 & 61.9 & 56.5 & 53.7 & 51.7 \\
    MMT & 71.3 & 66.0 & 62.7 & 58.2 & 55.5 & 53.5 \\
    MMT-cor & --- & 66.5 & 64.2 & 61.3 & 58.7 & 57.8 \\
    \bottomrule
    \end{tabular}
    }
    \caption{Results for word- and subword-level models trained and tested on noisy data. The word-level (w2w) results are used for comparison and are same as Table~\ref{tab:mix_noise}.}
    \label{tab:subword}
\end{table}

\section{Effect of $\lambda$}
\label{sec:lambda}
\begin{figure*}[htb]
\centering
  \includegraphics[width=.93\textwidth]{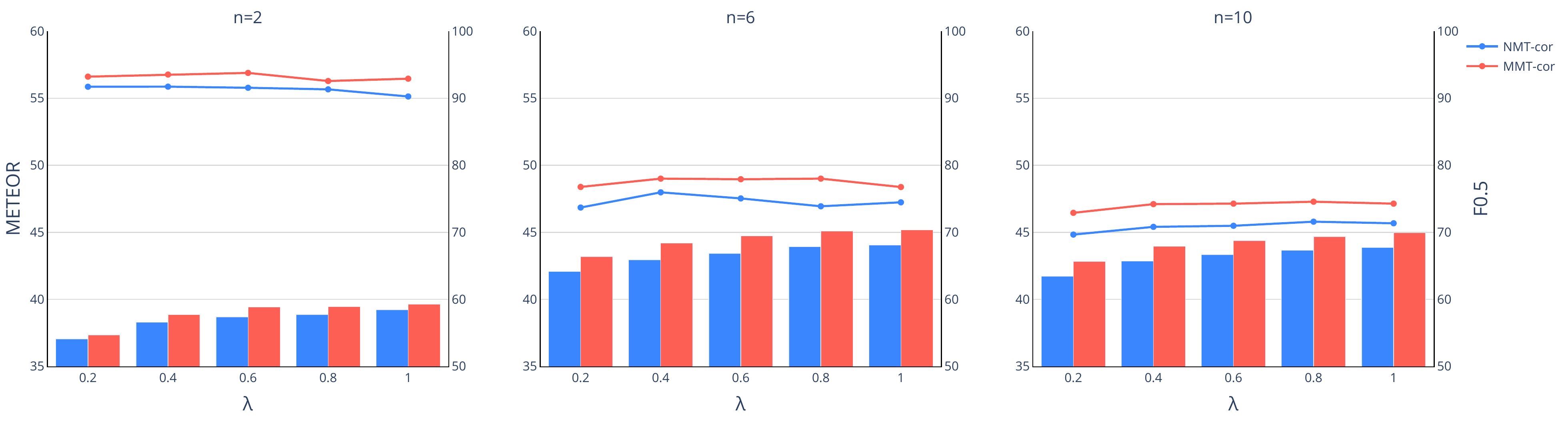}
  \caption{Effect of $\lambda$ on translation and error correction tasks. Lines: translation performance in METEOR score. Bars: error correction performance in $F_{0.5}$ score. The results are tested on MSCOCO2017 En-Fr data.}
  \label{fig:lambda_tune}
\end{figure*}

The value of $\lambda$ controls the weight of the error correction training for NMT-cor and MMT-cor models. This is thus an important hyper-parameter. We show the performance on translation and error correction tasks for different values of $\lambda$ in Figure~\ref{fig:lambda_tune}.

In terms of translation, the performance for both NMT-cor and MMT-cor models follows the same trend: the METEOR score first increases and then drops as $\lambda$ increases. This is reasonable since error correction is an auxiliary task, and a large weight for error correction task might harm models' ability to translate well. Nevertheless, the optimal $\lambda$ value is different for different levels of noise. Higher values of $\lambda$ help translating noisier texts. Regarding error correction, the increase of $\lambda$ always leads to better performance.

\section{Results with GRU Models}
\label{sec:gru_results}
In Table \ref{tab:mix_noise_gru}, we present the results for GRU models trained and tested on the noisy data. Similar to Transformer models, GRU models also benefits from multimodality and error correction training, and the improvement is larger on  noisier data.

In Table \ref{tab:clean_drop_gru}, the performance drop for GRU models on clean data is presented. Both MMT and MMT-cor shows lower drop than the NMT baseline, confirming that the improved robustness on noisy data does not sacrifice for the ability to translate clean data.
\begin{table*}[htb]
    \centering
    \begin{tabular}{l|c c c c c c | c c c c c c}
    \toprule
     & \multicolumn{6}{c}{\textbf{flickr2017}} & \multicolumn{6}{|c}{\textbf{mscoco2017}}  \\
     & clean & $n$=1 & $n$=2 & $n$=4 & $n$=6 & $n$=10 & clean & $n$=1 & $n$=2 & $n$=4 & $n$=6 & $n$=10 \\
    \midrule
    \multicolumn{11}{@{}l}{\textit{en-fr}} \\
    NMT & 70.3 & 64.9 & 61.1 & 55.9 & 53.0 & 50.7 & 64.7 & 59.5 & 55.4 & 49.2 & 45.8 & 43.7 \\
    MMT & 70.9 & 65.3 & 61.9 & 57.5 & 54.5 & 52.0 & 65.2 & 59.7 & 56.5 & 50.9 & 47.4 & 45.2 \\
    NMT-cor & --- & 65.2 & 61.8 & 57.3 & 54.6 & 53.1 & --- & 59.8 & 55.9 & 50.6 & 48.0 & 45.8 \\
    MMT-cor & --- & \textbf{65.4} & \textbf{62.5} & \textbf{58.4} & \textbf{56.0} & \textbf{54.3} & --- & \textbf{60.3} & \textbf{56.6} & \textbf{51.5} & \textbf{49.0} & \textbf{47.3} \\
    
    \midrule
    \multicolumn{11}{@{}l}{\textit{en-de}} \\
    NMT & 52.3 & 48.0 & 45.3 & 41.5 & 39.7 & 36.8 & 47.3 & 43.5 & 40.9 & 36.2 & 34.6 & 30.7 \\
    MMT & 52.5 & 48.5 & 45.9 & 42.5 & 40.6 & 39.4 & 47.4 & 43.9 & 41.3 & 37.7 & 35.3 & 33.7 \\
    NMT-cor & --- & \textbf{48.6} & 46.3 & 43.1 & 40.7 & 39.1 & --- & 44.1 & 41.7 & 37.4 & 35.5 & 33.3 \\
    MMT-cor & --- & 48.5 & \textbf{46.7} & \textbf{44.0} & \textbf{42.6} & \textbf{41.3} & --- & \textbf{44.3} & \textbf{42.0} & \textbf{39.0} & \textbf{37.3} & \textbf{35.6} \\
    
    \bottomrule
    \end{tabular}
    \caption{Results for GRU models trained and tested on different levels of noisy data. The train and test data are injected with the same proportion of noise.}
    \label{tab:mix_noise_gru}
\end{table*}

\begin{table}[H]
    \centering
    \begin{tabular}{l | c c c c c}
    \toprule
         $n$ = & 1 & 2 & 4 & 6 & 10 \\
    \midrule
        NMT & $\downarrow$0.4 & $\downarrow$0.5 & $\downarrow$1.5 &	$\downarrow$2.3 & $\downarrow$3.1 \\
        MMT & $\downarrow$0.2 & $\downarrow$0.9 & $\downarrow$1.3 &	$\downarrow$2.2 & $\downarrow$2.4 \\
        MMT-cor & $\downarrow$0.2 & $\downarrow$0.6 & $\downarrow$1.6 &	$\downarrow$1.9 & $\downarrow$2.7 \\
        
        
    \bottomrule
    \end{tabular}
    \caption{Performance drop (the lower the better) on the clean Flickr2017 En-Fr test set when GRU models are \textit{trained on mixed data} but \textit{tested on clean data}.}
    \label{tab:clean_drop_gru}
\end{table}

These results with GRU models further confirm that both multimodality and error correction training  improves translation robustness and can generalise to different models.

\section{Performance Drop on Clean Texts (Trained on Fully Noisy Data)}
\label{sec:train_noise_test_clean}
In Table \ref{tab:clean_drop_full_noise}, we present the performance drop on clean texts for models trained on fully noisy data. The drop on clean texts is not obvious for models trained with smaller \textit{$n$} while as \textit{$n$} becomes large, all three models suffers from a significant perform degradation. The results indicates that the proportion of noise in the training data is an important factor for robustness. However, to a lesser extent, the benefit from visual context and error correction training still holds on the clean test set, which indicates that the two methods do not simply trade off the performance on clean and noisy texts.
\begin{table}[H]
    \centering
    \begin{tabular}{l | c c c c c}
    \toprule
         $n$ = & 1 & 2 & 4 & 6 & 10 \\
    \midrule
        NMT & $\downarrow$1.5 & $\downarrow$2.4 & $\downarrow$5.2 &	$\downarrow$9.3 & $\downarrow$19.8 \\
        MMT & $\downarrow$0.7 & $\downarrow$1.9 & $\downarrow$4.5 &	$\downarrow$8.6 & $\downarrow$18.1 \\
        MMT-cor & $\downarrow$0.8 & $\downarrow$1.7 & $\downarrow$4.4 &	$\downarrow$7.7 & $\downarrow$15.5 \\
     
    \bottomrule
    \end{tabular}
    \caption{Performance drop on the clean Flickr2017 En-Fr test set for models trained on completely noisy data, compared to baseline NMT model trained on clean data.}
    \label{tab:clean_drop_full_noise}
\end{table}

\section{Semantic Similarity}
\label{sec:similarity}
To study the effect of error correction training on the shared encoder, we conduct a semantic similarity evaluation for models w/o error correction training. For that, we extract the hidden states from the last encoder layer for each sentence and measure the average cosine similarity over all words between noisy sentences and their clean counterparts. The similarity is computed as:
\begin{equation}
    Sim(\mathbf{x'}, \mathbf{x}) = \frac{1}{k}\sum_{i=1}^{k}\frac{\mathbf{h'}_i\cdot \mathbf{h}_i}{\lVert\mathbf{h'}_i\rVert\cdot\lVert\mathbf{h}_i\rVert}
\end{equation}
where $\mathbf{x'}=[x'_1, x'_2, ..., x'_k]$ represents the noisy sentence, $\mathbf{x}=[x_1, x_2, ..., x_k]$ represents the clean sentence, and $\mathbf{h'}_i$ and $\mathbf{h}_i$ represent the hidden state vectors for the $i$-th word in the noisy/clean sentences respectively.

Results are presented in Table \ref{tab:cosine}. Models applied with the error correction training achieve higher similarity between the clean and noisy hidden representations, suggesting that the error correction task helps learn a noise-invariant encoder representation. It is also interesting that visual features can slightly improve the similarity. The reason might be that the model learns alignments for both (image, clean text) and (image, noisy words). Therefore, the image might act as a bridge connecting the clean and noisy texts.
\begin{table}[H]
    \centering
        \begin{tabular}{l | c c c c c}
        \toprule
           {\textit{$n$}=} & 1 & 2 & 4 & 6 & 10 \\
        \midrule
            
            NMT & .980 & .964 & .935 & .915 & .902 \\
            NMT-cor & .984 & .970 & .946 & .928 & .918 \\
        \midrule
            
            MMT & .982 & .968 & .940 & .922 & .911 \\
            MMT-cor & .986 & .973 & .952 & .937 & .926 \\
        \bottomrule
        \end{tabular}
    \caption{Cosine similarity between the hidden representations for noisy and clean sentences. All models are trained with $\textit{n}$=4 and tested on Flickr2017 En-Fr.}
    \label{tab:cosine}
\end{table}

\section{More Qualitative Examples}
\label{sec:qualitative_examples}
\begin{figure*}[htb]
\centering
  \includegraphics[width=\textwidth]{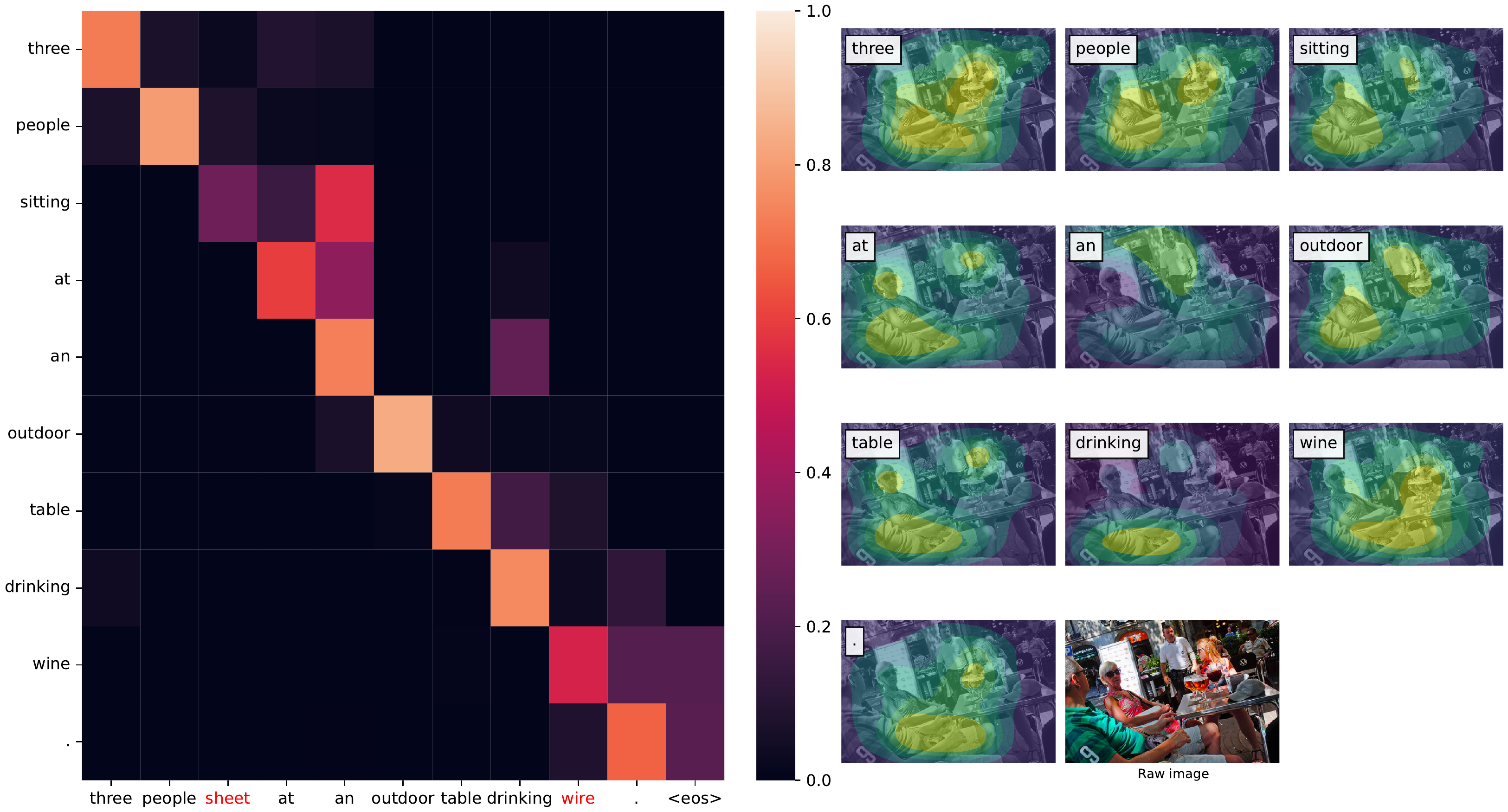}
  \caption{Attention map of the MMT-cor system on input texts and visual features when generating the error correction from noisy input with the correction decoder.}
  \label{fig:cor_attention}
\end{figure*}

In the appendix we provide some qualitative examples of translation (Table \ref{fig:more_translation_examples}) and error correction (Table \ref{fig:more_error_correction_examples}, and Figure \ref{fig:cor_attention}).
\begin{table*}[htb]
\renewcommand{\arraystretch}{1.0}
\centering
\begin{tabular}{cl@{}}
\toprule
\MR{7}{*}{\includegraphics[height=2.5cm]{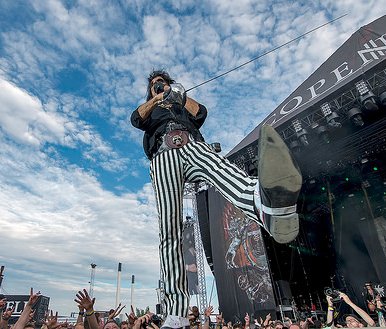}} &
\T{SRC}: a man in pinstripe pants is performing a concert . \\
& \T{NSY}: a man in pinstripe pants is [perforning] a [concett] . \\
& \T{NMT}: un homme en pantalon \false{beige} \false{prend} un \true{concert} . \\
& \phantom{\T{NMT}:} \I{(a man in beige pants is taking a concert.)} \\
& \T{MMT}: un homme en pantalon rayé \true{fait} un \true{concert} . \\
& \phantom{\T{MMT}:} \I{(a man in pinstripe pants is performing a concert.)} \\
& \T{MMT\textsubscript{cor}}: un homme en pantalon rayé \true{fait} un \true{concert} . \\
& \T{REF}: un homme en pantalon rayé fait un concert . \\
\midrule
\MR{9}{*}{\includegraphics[height=2.5cm]{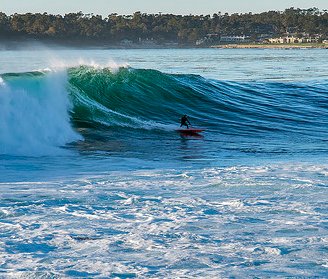}} &
\T{SRC}: a surfer rides a big wave . \\
& \T{NSY}: a surfer [ridez] a big [qave] . \\
& \T{NMT}: un surfeur prend une grosse vague . \\
& \phantom{\T{NMT}:} \I{(a surfer takes a big wave.)} \\
& \T{MMT}:un surfeur avec une grosse vague . \\
& \phantom{\T{NMT}:} \I{(a surfer with a big wave.)} \\
& \T{MMT\textsubscript{cor}}:un surfeur surfe une grosse vague . \\
& \T{REF}: un surfeur surfe une grosse vague . \\
\bottomrule
\end{tabular}
\caption{Translation examples generated by NMT, MMT and MMT-cor models. Noise is indicated by the words in square brackets. Underlined and bold words highlight the bad and good lexical choices, respectively.}
\label{fig:more_translation_examples}
\end{table*}

\begin{table*}[htb]
\renewcommand{\arraystretch}{1.0}
\centering
\begin{tabular}{cl@{}}
\toprule
\MR{6}{*}{\includegraphics[height=2.5cm]{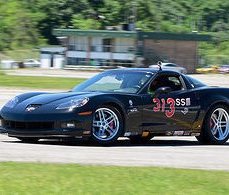}} &
\\
& \T{SRC}: there is a black car on a race track . \\
& \T{NSY}: there is a [blafk] [cat] on a race track . \\
& \T{COR-NMT}: there is a \true{black} \false{cat} on a race track . \\
& \T{COR-MMT}: there is a \true{black} \true{car} on a race track . \\
\\
\midrule
\MR{6}{*}{\includegraphics[height=2.5cm]{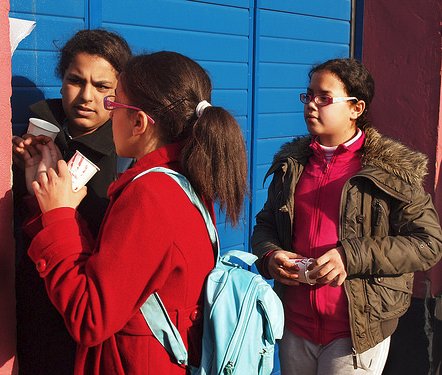}} & \\
& \T{SRC}: three girls with paper cups engage in conversation . \\
& \T{NSY}: [ree] girls with [pape] cups engage in conversation . \\
& \T{COR-NMT}: \true{three} girls with paper cups participate in conversation . \\
& \T{COR-MMT}: \true{three} girls with paper cups \true{engage} in conversation . \\
\\
\midrule
\MR{6}{*}{\includegraphics[height=2.5cm]{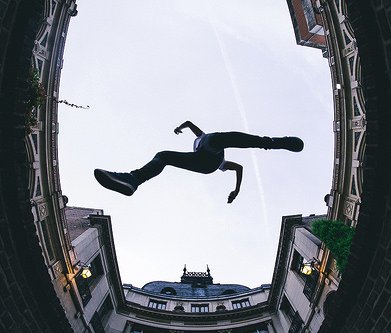}} & \\
& \T{SRC}: a person is leaping between two buildings . \\
& \T{NSY}: a [persson] is leaping between [tew] [building's] . \\
& \T{COR-NMT}: a \true{person} is \false{sleeping} between \true{two buildings} . \\
& \T{COR-MMT}: a \true{person} is \true{leaping} between \true{two buildings} . \\
\\
\bottomrule
\end{tabular}
\caption{Correction examples generated by NMT-cor and MMT-cor models. Noise is indicated by the words in square brackets. Underlined and bold words highlight the bad and good lexical choices, respectively.}
\label{fig:more_error_correction_examples}
\end{table*}

\end{document}